\documentclass{article}
\usepackage{ijcai21}

\usepackage{times}
\usepackage{soul}
\usepackage{url}
\usepackage[hidelinks]{hyperref}
\usepackage[utf8]{inputenc}
\usepackage[small]{caption}
\usepackage{graphicx}
\usepackage{amsmath}
\usepackage{amsthm}
\usepackage{booktabs}
\usepackage{algorithm}
\usepackage{algorithmic}
\title{Hardware-friendly Deep Learning by Network Quantization and Binarization}

\author{
Haotong Qin$^1$
\affiliations
$^1$Shen Yuan Honors College, Beihang University\\
\emails
qinhaotong@nlsde.buaa.edu.cn
}

\begin{document}

\maketitle

\begin{abstract}
  Quantization is emerging as an efficient approach to promote hardware-friendly deep learning and run deep neural networks on resource-limited hardware. However, it still causes a significant decrease to the network in accuracy. We summarize challenges of quantization into two categories: \textit{Quantization for Diverse Architectures} and \textit{Quantization on Complex Scenes}. Our studies focus mainly on applying quantization on various architectures and scenes and pushing the limit of quantization to extremely compress and accelerate networks. The comprehensive research on quantization will achieve more powerful, more efficient, and more flexible hardware-friendly deep learning, and make it better suited to more real-world applications.
\end{abstract}

\section{Background}
With the continuous development of deep learning, deep neural networks (DNNs) have made significant progress in various fields, such as computer vision, natural language processing, and speech recognition. Owing to the deep structure with a number of layers and millions of parameters, the DNNs enjoy strong learning capacity, and thus usually achieve satisfactory performance.
For example, the most advanced language model GPT-3 contains about 175 billion 32-bit floating-point parameters and requires about 700GB of memory to be stored.
This fact makes the DNNs heavily rely on high-performance hardware such as GPU, while in real-world applications, only the devices (e.g., mobile and embedded devices) with limited resources are available. 



\section{Quantization and Binarization}
Quantization and binarization are emerged as efficient approaches to compress and accelerate the neural network. They quantize or binarize the FP32 parameters in neural network models to lower bit-width, such as 1-8 bit, to compress the models, and also accelerate models by replacing FP32 Multiply–Accumulate (MAC) operations with efficient integer or bitwise operations.

For multi-bit uniform quantization, given the bit-width $b$ and the FP32 activation/weight $x$ following in the range($l$, $u$), the quantization process of uniform quantization can be defined as:
\begin{equation}
x_Q=\operatorname{round}\left(\frac{x}{\Delta}\right),
\end{equation}
where the original range $(l, u)$ is divided into $2^{b}-1$ intervals $\mathcal{P}_{i}, i \in\left(0,1, \ldots, 2^{b}-1\right)$, and $\Delta=\frac{u-l}{2^{b}-1}$ is the interval length.
And the dequantization process is:
\begin{equation}
x_{\text{out}}=x_Q \Delta.
\end{equation}
For 1-bit binarization, the activation/weight is binarized to either -1 or +1, usually using the binary function:
\begin{equation}
x_\text{out}=\operatorname{sign}(x)=\left\{\begin{array}{ll}
+1, & \text { if } x \geq 0 \\
-1, & \text { otherwise }
\end{array}\right.
\end{equation}
The binarization can be considered as a special case of quantization aiming to extremely compress the bit-width.
The backward pass is modeled as a "straight through estimator" (STE). Specifically,
$$
\delta_{\text{out}}=\delta_{\text{in}} I_{x\in (l, u)},
$$
where $\delta_\text{in}=\frac{\partial L}{\partial w_\text{out}}$ is the backpropagation error of the loss with respect to the quantizer output.

Benefited for the compact quantized parameters, the model size of quantized networks can be significantly compressed with FP32 counterparts. The acceleration capability of a quantized network has been proved from both theoretical and practical aspects. The binarized network based on 1-bit representation enjoys the compressed storage and fast inference speed, e.g., the binarized network enjoys up to $38\times$ speedups and $89\times$ energy efficiency over existing frameworks on mobile GPUs~\cite{chen2020phonebit}. However, it meanwhile suffers from performance degradation.
Compared with other methods, quantization and binarization approaches pay more attention to obtain the compact low-bit parameters with better representation rather than optimize the network architectures, thereby these approaches enjoy more versatility.

\section{Our Studies}

We summarize challenges of quantization into two categories: 

\noindent (1) \textbf{Quantization for Diverse Architectures}. For processing various types of data (e.g., 2D image, 3D point clouds, and text data), more neural networks with diverse architectures are proposed. But approaches suitable for a certain architecture depend on the data/feature and structure attributes.

\noindent (2) \textbf{Quantization on Complex Scenes}. Most advanced quantization approaches rely on the expensive retraining process and real training data, while the requirements for quantization also exist on many complex scenes, such as data-free, resource-limited, time-limited scenes.

Therefore, our studies focus mainly on applying quantization on various architectures and scenes and pushing the limit of quantization to extremely compress and accelerate networks.
Our survey paper presents a comprehensive survey of binarization approaches and also investigates other practical aspects of binary neural networks such as the hardware-friendly design and the training tricks~\cite{qin2020binary}.

\subsection{Quantization-Aware Training}

Quantization-aware training is an effective approach to reduce the degradation in model accuracy caused by quantization. Therefore, we studied that obtaining accurate neural networks with diverse architectures (CNNs, PointNet, etc.) by this approach.

\subsubsection{Binarized Convolutional Neural Networks}

Through enjoys extreme compact binarized parameters and efficient bitwise operations, there is a noticeable performance gap between the binarized model and the FP32 one that prevents binarization to be practical.
Our empirical study indicates that the quantization brings information loss in both forward and backward propagation, which is the bottleneck of training accurate binary neural networks~\cite{qin2020forward}.

To address the issues, we propose an Information Retention Network (IR-Net) to retain the information that consists in the forward activations and backward gradients. IR-Net mainly relies on two technical contributions, Libra Parameter Binarization and Error Decay Estimator. We are the first to investigate both forward and backward processes of binary networks from the unified information perspective, which provides new insight into the mechanism of network binarization. 

\subsubsection{Binarized PointNet for 3D Point Clouds}

To alleviate the resource constraint for real-time point cloud applications that run on edge devices, we further study to apply binarization to neural networks for deep learning on point clouds.
We discover that the immense performance drop of binarized models for point clouds mainly stems from two challenges: aggregation-induced feature homogenization that leads to a degradation of information entropy, and scale distortion that hinders optimization and invalidates scale-sensitive structures~\cite{qin2021bipointnet}.

We present BiPointNet, the first model binarization approach for efficient deep learning on point clouds. With theoretical justifications and in-depth analysis, our BiPointNet introduces Entropy-Maximizing Aggregation to modulate the distribution before aggregation for the maximum information entropy, and Layer-wise Scale Recovery to efficiently restore feature representation capacity.
BiPointNet gives an impressive $14.7\times$ speedup and $18.9\times$ storage saving on real-world resource-constrained devices.

\subsection{Data-Free Quantization}

Recently, data-free quantization has been widely studied as a practical and promising solution. It synthesizes data for calibrating the quantized model according to the batch normalization statistics of FP32 ones and significantly relieves the heavy dependency on real training data in traditional quantization methods. We find that in practice, the synthetic data identically constrained by BN statistics suffers serious homogenization at both distribution level and sample level and further causes a significant performance drop of the quantized model~\cite{zhang2021diversifying}.

We propose Diverse Sample Generation (DSG) scheme to mitigate the adverse effects caused by homogenization.
Our DSG obtains significant improvements over various network architectures and quantization methods, especially when quantized to lower bits, and models calibrated with synthetic data perform close to those calibrated with real data and even outperform them on W4A4.

\section{Discussion}

Our long-term research goal is to enable state-of-the-art neural network models to be deployed on resource-limited hardware, which includes the compression and acceleration for multiple architectures and scenes, and the flexible and efficient deployment on multiple hardware.
Now we are attempting to apply quantization and binarization to more architectures, such as transformer-based BERT models, and improve their performance over extreme low bit-width.
In fact, prior studies have proved that there usually exists large redundancy in the deep structure.
Therefore, the networks quantized by existing quantization approaches are far from reaching the limit of their performance, and the quantization and binarization are still worthy of continuous attention and study.

Accurate quantization and binarization approaches can significantly expand the deployment scenes of advanced neural networks. Moreover, under the background of the explosive growth of computation requirement of state-of-the-art neural networks, this study will help relieve the inequality-producing effects of AI, and let more individuals or small and medium-sized enterprises enjoy the advanced neural network with low cost.

{\small
\bibliographystyle{named}
\bibliography{ijcai21}
}

\end{document}